\documentclass[runningheads,a4paper]{llncs}
\usepackage[OT1]{fontenc}
\usepackage{makeidx}  % allows for indexgeneration
\usepackage{amsmath}
\usepackage{array}
\usepackage{multirow}
\usepackage{latexsym}
\usepackage{CJK}
\usepackage{algorithm}
\usepackage{algorithmic}
\usepackage{color}
\usepackage{array}
\usepackage{ulem}%
\usepackage{amssymb}
\usepackage{graphicx}

\usepackage{url}
\urldef{\mails}\path|{xwshi, hhy63, pjian, guoyuhang, wxchi, tangyk}@bit.edu.cn|
\newcommand{\keywords}[1]{\par\addvspace\baselineskip
\noindent\keywordname\enspace\ignorespaces#1}
\begin{document}
\mainmatter  % start of an individual contribution

\title{Neural Chinese Word Segmentation as \\Sequence to Sequence Translation}

\author{ Xuewen Shi\and Heyan Huang\and
Ping Jian%\thanks{Corresponding author}
\and Yuhang Guo\and \\
Xiaochi Wei\and Yikun Tang}
\authorrunning{X. Shi et al.}
\institute{
Beijing Engineering Research Center of High Volume Language Information Processing
and Cloud Computing Applications, \\
School of Computer Science and Technology, \\
Beijing Institute of Technology, Beijing 100081, China\\
\mails\\
}

\toctitle{Lecture Notes in Computer Science}
\tocauthor{Authors' Instructions}
\maketitle

\begin{abstract}
Recently, Chinese word segmentation (CWS) methods using neural networks have made impressive progress.
Most of them regard the CWS as a sequence labeling problem which construct models based on local features rather than considering global information of input sequence.
In this paper, we cast the CWS as a sequence translation problem and
propose a novel sequence-to-sequence CWS model with an attention-based encoder-decoder framework.
The model captures the global information from the input and directly outputs the segmented sequence.
It can also tackle other NLP tasks with CWS jointly in an end-to-end mode.
Experiments on Weibo, PKU and MSRA benchmark datasets show that
our approach has achieved competitive performances compared with state-of-the-art methods.
Meanwhile, we successfully applied our proposed model to jointly learning CWS and Chinese spelling correction,
which demonstrates its applicability of multi-task fusion.
\end{abstract}
\section{Introduction}
Chinese word segmentation (CWS) is an important step for most Chinese natural language processing (NLP) tasks, since Chinese is usually written without explicit word delimiters.
The most popular approaches treat CWS as a sequence labelling problem~\cite{Xue:2003,Peng:2004} which can be handled with supervised learning algorithms, e.g. Conditional Random Fields~\cite{lafferty:2001,Peng:2004,zhao2010unified,sun2014feature}.
However the performance of these methods heavily depends on the design of handcrafted features.

Recently, neural networks for CWS have gained much attention as they are capable of learning features automatically.
Zheng et al.~\cite{zheng2013deep} adapted word embedding and the neural sequence labelling architecture~\cite{collobert2011natural} for CWS.
Chen et al.~\cite{chen2015gated} proposed gated recursive neural networks to model the combinations of context characters.
Chen et al.~\cite{chen2015long} introduced Long Short-Term Memory (LSTM) into neural CWS models to capture the potential long-distance dependencies.
The aforementioned methods predict labels of each character in the order of
the sequence by considering context features within a fixed-sized window and limited tagging history~\cite{cai2016neural}.
In order to eliminate the restrictions of previous approaches, we cast the CWS as a sequence-to-sequence translation task.

The sequence-to-sequence framework has successful applications in machine translation~\cite{sutskever2014sequence,bahdanau2014neural,wu2016google,gehring2017convs2s},
which mainly benefits from
(i) distributed representations of global input context information,
(ii) the memory of outputs dependencies among continuous timesteps and
(iii) the flexibilities of model fusion and transfer.

In this paper, we conduct sequence-to-sequence CWS under an attention-based recurrent neural network (RNN) encoder-decoder framework.
The encoder captures the whole bidirectional input information without context window limitations.
The attention based decoder directly outputs the segmented sequence by simultaneously considering the global input context information and the dependencies of previous outputs.
Formally, given an input characters sequence $\mathbf{x}$ with $T$ words i.e. $\mathbf{x}=(x_1,x_2,...,x_{T_x})$, our model directly generates an output sequence $\mathbf{y}=(y_1,y_2,...,y_T)$ with segmentation tags inside. For example, given a Chinese sentence ``\begin{CJK*}{UTF8}{gbsn} 我爱夏天\end{CJK*}" (I love summer) , the input
$\mathbf{x}=($
\begin{CJK*}{UTF8}{gbsn}我\end{CJK*},
\begin{CJK*}{UTF8}{gbsn}爱\end{CJK*},
\begin{CJK*}{UTF8}{gbsn}夏\end{CJK*},
\begin{CJK*}{UTF8}{gbsn}天\end{CJK*}
$)$ and the output $\mathbf{y}=($
\begin{CJK*}{UTF8}{gbsn}我\end{CJK*},
$<$/s$>$,
\begin{CJK*}{UTF8}{gbsn}爱\end{CJK*},
$<$/s$>$,
\begin{CJK*}{UTF8}{gbsn}夏\end{CJK*},
\begin{CJK*}{UTF8}{gbsn}天\end{CJK*}
$)$ where the symbol `$<$/s$>$' denotes the segmentation tag.
In the post-processing step, we replace `$<$/s$>$' with word delimiters and join the characters sequence into a sentence as $\mathbf{s}=$
``\begin{CJK*}{UTF8}{gbsn}我 爱 夏天\end{CJK*}".

In addition, considering that the sequence-to-sequence CWS is an end-to-end process of natural language generation,
it has the capacity of jointly learning with other NLP tasks.
In this paper, we have successfully applied our proposed method to jointly learning CWS and Chinese spelling correction (CSC) in an end-to-end mode,
which demonstrates the applicability of the sequence-to-sequence CWS framework.

We evaluate our model on three benchmark datasets, Weibo, PKU and MSRA.
The experimental results show that the model achieves competitive performances compared with state-of-the-art methods.

The main contributions of this paper can be summarized as follows:
\begin{itemize}
\item[$\bullet$]
We first treat CWS as a sequence-to-sequence translation task and introduce the attention-based encoder-decoder framework into CWS.
The encoder-decoder captures the whole bidirectional input information without context window limitations
and directly outputs the segmented sequence by simultaneously considering the dependencies of previous outputs and the input information.
\item[$\bullet$]
We let our sequence-to-sequence CWS model simultaneously tackle other NLP tasks, e.g., CSC,
in an end-to-end mode, and we well validate its applicability in our experiments.
\item[$\bullet$]
We propose a post-editing method based on longest common subsequence (LCS)~\cite{lin2004automatic}
to deal with the probable translation errors of our CWS system.
 This method solves the problem of missing information in the translation process and improves the experiment results.
%\item We pre-train our model using pseudo data with different distributions from the training datasets to alleviate the problem of insufficient data.
\end{itemize}

\section{Method}

\subsection{Attention based RNN Encoder-decoder Framework for CWS}
\label{sec:model}

Our approach uses the attention based RNN encoder-decoder architecture called RNNsearch~\cite{bahdanau2014neural}.
From a probabilistic perspective, our method is equivalent to finding a character sequence $\mathbf{y}$ with segmentation tags inside via maximizing the conditional probability of $\mathbf{y}$ given a input character sequence $\mathbf{x}$, i.e., $argmax_{y}p(\mathbf{y}|\mathbf{x})$.

\begin{figure}[t]
\centering
\includegraphics[width=0.68\textwidth]{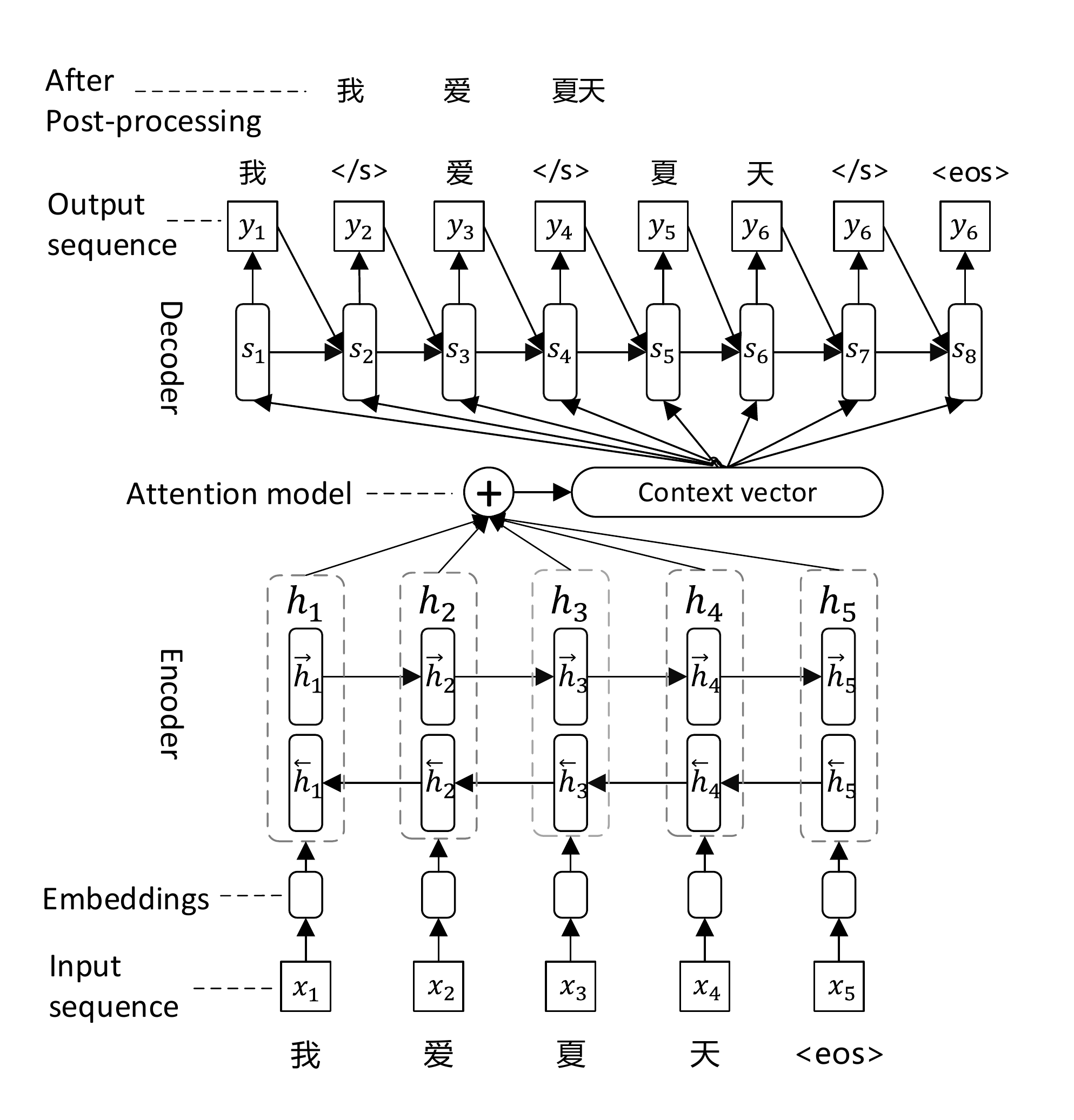}
\caption{ Illustration of the presented model for CWS. The tag `$<$eos$>$' refers to the end of the sequence. }
\label{inputoutput}
\end{figure}

The model contains (i) an bidirectional RNN encoder to maps the input $\mathbf{x}=(x_1,x_2,...,x_{T_x})$ into a sequence of annotations $(h_1,h_2,...,h_{T_x})$, and (ii) an attention based RNN decoder to generate the output sequence $\mathbf{y}=(y_1,y_2,...,y_T)$. Fig. \ref{inputoutput} gives an illustration of the model architecture.

\subsection{bidirectional RNN Encoder}
The bidirectional RNN encoder consists of forward and backward RNNs. The forward RNN $\overrightarrow{f}$ reads the input sequence in the order of (from $x_1$ to $x_{T_x}$) and calculates the sequence $(\overrightarrow{h}_1,\overrightarrow{h}_2,...,\overrightarrow{h}_{T_x})$, while the backward RNN $\overleftarrow{f}$ reads the input sequence in the reverse order of (from $x_{T_x}$ to $x_1$) and calculates the sequence $(\overleftarrow{h}_1,\overleftarrow{h}_2,...,\overleftarrow{h}_{T_x})$. Finally, the annotation $h_j$ for each $x_j$ is obtained by
$h_j=
\begin{bmatrix}
 {\overrightarrow{h}_j}^T; {\overleftarrow{h}_j}^T
\end{bmatrix}^T\textrm{.} \quad
$
\subsection{Attention-based RNN Decoder}

The attention-based RNN decoder estimates the conditional probability $p(\mathbf{y}|\mathbf{x})$ as
\begin{equation}
\label{eq::1}
p(\mathbf{y}|\mathbf{x})=
\prod_{t}^{T}p(y_t|{y_1,...,y_{t-1}},\mathbf{x})
\textrm{.}
\end{equation}
In Eq.(\ref{eq::1}), each conditional probability is defined as:
\begin{equation}
\label{eq::2}
p(y_t|{y_1,...,y_{t-1}},\mathbf{x})=g(y_{t-1},s_t,c_t)\textrm{,}
\end{equation}
where $s_t$ is the RNN hidden state for time $t$ and computed by
\begin{equation}
\label{eq::3}
s_t=f(s_{t-1}, y_{t-1}, c_t)
\textrm{.}
\end{equation}

The $c_t$ in Eq.(\ref{eq::2}) and Eq.(\ref{eq::3}) is the context vector computed as a weighted sum of the annotations $(h_1,h_2,...,h_{T_x})$:
$$
c_t=\sum_{j=1}^{T_x}\alpha_{t,j}h_j
\textrm{.}
$$
The weight $\alpha_{t,j}$ is computed by:
$$
\alpha_{t,j}=\frac{exp(e_{t,j})}{\sum_{k=1}^{T_x}exp(e_{t,k})}
\textrm{,}
$$
where $e_{t,j}=a(s_{t-1},h_j)$, therein, $a(\boldmath{\cdot})$
is an attention model constructed with a feedforward neural network.

\subsection{Post-editing Method for Sequence-to-sequence CWS}

We found some negative outputs of our model caused by translation errors such as missing words and extra words.
The cause of the errors is mostly due to out-of-vocabulary or rare Chinese characters of input sequence.

Table~\ref{tab:postediting_example} shows an example with translation errors of our sequence-to-sequence CWS system.
The original input comes from the Weibo dataset (seen in Section~\ref{sec::datasets}).
The output missed three Japanese characters ``\begin{CJK*}{UTF8}{gbsn}極\end{CJK*}''(extreme), ``\begin{CJK*}{UTF8}{gbsn}の\end{CJK*}''(of) and ``\begin{CJK*}{UTF8}{gbsn} 親\end{CJK*}''(parent),
and introduced three extra characters ``UNK'' instead which means `unknown word' in the vocabulary.

\begin{table}[h]
  \caption{An example of translation errors in our CWS system and post-editing results.}
  \label{tab:postediting_example}
  \centering
  \renewcommand\arraystretch{1.1}
  \begin{tabular}{|p{2.0cm}<{\raggedleft} |p{9.5cm}<{\raggedright} |}
  \hline
  Original input &\begin{CJK*}{UTF8}{gbsn}岛国一超精分的小品《\uline{\textcolor{blue}{極}}道\uline{\textcolor{blue}{の親}}子》，看完之后我想说，为什么我没有这么“通情达理”的老爸呢？\end{CJK*} \\ \hline
  System output &\begin{CJK*}{UTF8}{gbsn}岛国 一 超 精分 的 小品 《 \uuline{\textcolor{red}{UNK}}道 \uuline{\textcolor{red}{UNK}} \uuline{\textcolor{red}{UNK}}子 》 ， 看完 之后 我 想 说 ， 为什么 我 没有 这么 “ 通情达理 ” 的 老爸 呢 ？\end{CJK*} \\ \hline
  After post-editing &\begin{CJK*}{UTF8}{gbsn}岛国 一 超 精分 的 小品 《 \uline{\textcolor{blue}{極}}道 \uline{\textcolor{blue}{の}} \uline{\textcolor{blue}{親}}子 》 ， 看完 之后 我 想 说 ， 为什么 我 没有 这么 “ 通情达理 ” 的 老爸 呢 ？\end{CJK*}\\\hline
  Gold standard &\begin{CJK*}{UTF8}{gbsn}岛国 一 超 精分 的 小品 《 \uline{\textcolor{blue}{極}}道 \uline{\textcolor{blue}{の}} \uline{\textcolor{blue}{親}}子 》 ， 看完 之后 我 想 说 ， 为什么 我 没有 这么 “ 通情达理 ” 的 老爸 呢 ？\end{CJK*}\\\hline
  \end{tabular}
\end{table}

\renewcommand{\algorithmicrequire}{ \textbf{Input:}}
\renewcommand{\algorithmicensure}{ \textbf{Output:}}
\begin{algorithm}[!h]
\caption{ Post-editing algorithm for our CWS system }
\label{alg:postediting}
\begin{algorithmic}
\REQUIRE ~~\\
The original input character sequence: $s_{ori}$;\\
The segmented word sequence with translation errors from sequence-to-sequence CWS system: $s_{seg}$;\\
\ENSURE ~~\\
\STATE Segmentation labels set: $L\gets\{B,M,E,S\}$;
%\label{ code:fram:definelabels }
\STATE Labeling characters in $s_{seg}$ with labels in $L$ gets $lab_{seg}$;
\STATE $length_{ori}\gets getLength(s_{ori})$, $length_{seg}\gets getLength(s_{seg})$
\IF{$length_{ori}\neq length_{seg}$}
\STATE Labeling characters in $s_{ori}$ with position labels;
%\label{ code:fram:locationlabels }
\STATE Extracting the longest common subsequences between $s_{ori}$ and $s_{seg}$ using longest common subsequence (LCS) algorithm: $s_{sub} = LCS(s_{seg}, s_{ori})$;
%\label{ code:fram:lcs }
\STATE Taking $s_{ori}$ as a reference, filling the missing characters in $s_{sub}$ and labeling them with label $X$;
%\label{ code:fram:fill }
\STATE Replacing label $X$ with labels in $L$ according to manually prepared rules;
%\label{ code::fram:replacex }
\ELSE
\STATE Labeling $s_{ori}$ according to $lab_{seg}$;
\ENDIF
\STATE Merging the characters in $s_{ori}$ into word sequence $s_{pe}$ according to their segmentation labels;
%\label{ code:fram:segmentationlabels }
\RETURN $s_{pe}$;
\end{algorithmic}
\end{algorithm}

Inspired by Lin and Och~\cite{lin2004automatic}, we proposed an LCS based post-editing algorithm\footnote{Executable source code is available at \\ \url{https://github.com/SourcecodeSharing/CWSpostediting}} (seen in Algorithm~\ref{alg:postediting}) to alleviate the negative impact to CWS.
In the algorithm, we define an extended word segmentation labels set $\{B,M,E,S,X\}$.
$\{B,M,E\}$ represent begin, middle, end of a multi-character segmentation respectively,
and $S$ represents a single character segmentation.
The additional label $X$ in $L$ can be seen as any other labels according to its context.
For example, given a CWS label sequence $(S,S,B,E,B,X,E)$, the $X$ should be transformed into label $M$ and
in the other case of $(S,X,B,E,B,M,E)$, the $X$ should be treated as label $S$.
The above transformation strategy can be based on handcraft rules or machine learning methods.
In this paper, we use the transformation rules written manually.
Table~\ref{tab:postediting_example} also gives an example of post-editing results.
%The differences of P, R and F scores between the sequence-to-sequence CWS results and post-editing results are shown in Table 4 and Table 5.

\section{Experiments}

\subsection{Datasets}
\label{sec::datasets}
We use three benchmark datasets, Weibo, PKU and MSRA, to evaluate our CWS model. Statistics of all datasets are shown in Table~\ref{tab:datasets}.

\textbf{Weibo}\footnote{All data and the program are available at \\\url{https://github.com/FudanNLP/NLPCC-WordSeg-Weibo}}:
this dataset is provided by NLPCC-ICCPOL 2016 shared task of Chinese word segmentation for Micro-blog Texts~\cite{qiu2016overview}.
The data are collected from Sina Weibo\footnote{\url{http://www.weibo.com}}.
Different with the popular used newswire dataset, the texts of the dataset are relatively informal and consists various topics.
Experimental results on this dataset are evaluated by eval.py scoring program$^1$, which calculates standard precision (P), recall (R) and F1-score (F) and weighted precision (P), recall (R) and F1-score (F)~\cite{qiu2016new}~\cite{qiu2016overview} simultaneously.

\textbf{PKU and MSRA}\footnote{All data and the program are available at \\\url{http://sighan.cs.uchicago.edu/bakeoff2005/}:}
these two datasets are provided by the second International Chinese Word Segmentation Bakeoff~\cite{emerson2005second}.
We found that the PKU dataset contains many long paragraphs consisting of multiple sentences, which has negative impacts on the training of the sequence translation models.
To solve this problem, we divide the long paragraphs in the PKU dataset into sentences.
Experiment results on those two datasets are evaluated by the standard Bakeoff scoring program$^3$, which calculates P, R and F scores.

\begin{table*}[t]
  \caption{Statistics of different datasets. The size of training/testing datasets are given in number of sentences (Sents), words (Words) and characters (Chars).}
  \label{tab:datasets}
  \centering
  \begin{tabular}{|p{1.3cm}<{\centering}|p{1.3cm}<{\centering}|p{1.5cm}<{\centering}|p{1.5cm}<{\centering}|p{1.1cm}<{\centering}|p{1.3cm}<{\centering}|p{1.3cm}<{\centering}|}
  \hline
  \multirow{2}{*}{Datasets} &
  \multicolumn{3}{c|}{Training} &
  \multicolumn{3}{c|}{Testing} \\
  \cline{2-7}
  &Sents &Words &Chars &Sents &Words &Chars \\
  \hline
  Weibo &20,135 &421,166 &688,743 &8,592 &187,877 &315,865 \\
  PKU &43,475 &1,109,947 &1,826,448  &4,261 &104,372 &172,733\\
  MSRA &86,924 &2,368,391 &4,050,469 &3,985 &106,873 &184,355\\
  \hline
  \end{tabular}
\end{table*}

\subsection{Model Setup and Pre-training}
We use the RNNsearch\footnote{Implementations are available at \url{https://github.com/lisa-groundhog/GroundHog}} model~\cite{bahdanau2014neural} to achieve our sequence-to-sequence CWS system.
The model is set with embedding size 620, 1000 hidden units and an alphabet with the size of 7190.
We also apply the Moses' phrase-based (Moses PB) statistical machine translation system~\cite{koehn2007moses} with 3-gram or 5-gram language model as sequence-to-sequence translation baseline systems.

Since our sequence-to-sequence CWS model contains large amount numbers (up to ten million) of free parameters,
it is much more likely to be overfitting when training on small datasets~\cite{srivastava2014dropout}.
In fact, we make an attempt to train the model on the benchmark datasets directly and get poor scores as shown in Table~\ref{tab::nopretrain}.
To deal with this problem, a large scale pseudo data is utilized to pre-train our model.
The Weibo, PKU and MSRA datasets are then used for fine-tuning.
To construct the pseudo data, we label the UN1.0~\cite{ziemski2016united} with
LTP\footnote{Available online at \url{https://github.com/HIT-SCIR/ltp}}~\cite{che2010ltp} Chinese segmentor.
The pseudo data contains 12,762,778 sentences in the training set and 4,000 sentences in the validation set and the testing set.
The testing set of the pseudo data is used to evaluate the pre-training performance of the model,
and the result P, R and F scores are \textbf{98.2}, \textbf{97.1} and \textbf{97.7} respectively w.r.t the LTP label as the ground truth.

\begin{table}[t]
  \caption{Experimental results on benchmark datasets w/o pre-training.}
  \label{tab::nopretrain}
  \centering
  \begin{tabular}{|p{2.9cm}<{\centering} |p{1.2cm}<{\centering} |p{1.2cm}<{\centering} |p{1.2cm}<{\centering}|}
  \hline
  Datasets &P &R &F \\ \hline
  Weibo &89.8 &89.5 &89.6 \\ \hline
  PKU &87.0 &88.6 &87.8 \\ \hline
  MSRA &95.1 &93.2 &94.1 \\ \hline
  %Pseudo Data &98.2 &97.1 &97.7 \\ \hline
  \end{tabular}
\end{table}

\subsection{CWS Experiment Results}

\begin{table*}[t]
  \caption{Experimental results on the CWS dataset of Weibo. The contents in parentheses represent the results of comparison with other systems.}
  \label{tab:weibo}
  \renewcommand\arraystretch{1.1}
  \centering
  \begin{tabular}{|c|m{4.7cm}<{\centering}|m{0.9cm}<{\centering}|m{0.9cm}<{\centering}|m{0.9cm}<{\centering}|m{0.9cm}<{\centering}|m{0.9cm}<{\centering}|m{0.9cm}<{\centering}|}
  \hline
  \multirow{2}{*}{Groups} &
  \multirow{2}{*}{Models} &
  \multicolumn{3}{c|}{ Standard Scores } &
  \multicolumn{3}{c|}{ Weighted Scores } \\
  \cline{3-8}
  & &P &R &F &P &R &F\\
  \hline
  A &LTP~\cite{che2010ltp} &83.98 &90.46 &87.09 &69.69 &80.43 &74.68 \\\hline
  \hline
  \multirow{3}{*}{B~\cite{qiu2016overview}}
  &S1 &94.13 &94.69 &94.41 &79.29 &81.62 &80.44 \\\cline{2-8}
  &S2 &94.21 &95.31 &94.76 &78.18 &81.81 &79.96 \\\cline{2-8}
  &S3 &94.36 &95.15 &94.75 &78.34 &81.34 &79.81 \\\cline{2-8}
  &S4 &93.98 &94.78 &94.38 &78.43 &81.20 &79.79 \\\cline{2-8}
  &S5 &93.93 &94.80 &94.37 &76.24 &79.32 &77.75 \\\cline{2-8}
  &S6 &93.90 &94.42 &94.16 &75.95 &78.20 &77.06 \\\cline{2-8}
  &S7 &93.82 &94.60 &94.21 &75.08 &77.91 &76.47 \\\cline{2-8}
  &S8 &93.74 &94.31 &94.03 &74.90 &77.14 &76.00 \\\cline{2-8}
  &S9 &92.89 &93.65 &93.27 &71.25 &73.92 &72.56 \\\hline
  \hline
  \multirow{7}{*}{M} &Moses PB w/ 3-gram LM &92.42 &92.26 &92.34 &76.74 &77.23 &76.98 \\\cline{2-8}
  &Moses PB w/ 5-gram LM &92.37 &92.26 &92.31 &76.58 &77.25 &76.91 \\\cline{2-8}
  &RNNsearch w/o fine-tuning &86.10 &88.82 &87.44 &68.88 &75.20 &71.90 \\\cline{2-8}
  &RNNsearch &92.09 &92.79 &92.44 &75.00 &78.27 &76.60 \\\cline{2-8}
  %&RNNsearch &92.09 &92.79 &92.44 &~75.00\par($>$S8) &~78.27\par($>$S6) &~76.60\par($>$S7)\\\cline{2-8}
  &RNNsearch w/ post-editing  &~93.48\par($>$S9) &~94.60\par($>$S6) &~94.04\par($>$S8) &~76.30\par($>$S5) &~79.99\par($>$S5) &~78.11\par($>$S5)\\
  \hline
  \end{tabular}
\end{table*}

\textbf{Weibo:}
for Weibo dataset, we compare our models with two groups of previous works on CWS as shown in Table~\ref{tab:weibo}.
The LTP~\cite{che2010ltp} in group A is a general CWS tool which we use to label pseudo data.
S1 to S8 in Group B are submitted systems results of NLPCC-ICCPOL 2016 shared task of Chinese word segmentation for Micro-blog Texts~\cite{qiu2016overview}.
Our works are shown in Group M.
Since the testing set of Weibo dataset has many out-of-vocabulary (OOV) words,
our post-editing method shows its effective for enhancing our CWS results for its abilities to recall missing words and replace extra words.

\textbf{PKU and MSRA:}
for the two popular benchmark datasets, PKU and MSRA, we compare our model with three groups of previous models on CWS task as shown in Table~\ref{tab:pkumsra}.
The LTP~\cite{che2010ltp} in group A is same as Table~\ref{tab:weibo}.
Group B presents a series of published results of previous neural CWS models with pre-trained character embeddings.
The work proposed by Zhang et al.~\cite{zhang2013exploring} in group C is one of the state-of-the-art methods.
Our post-editing method dose not significantly enhance the CWS results for PKU and MSRA datasets comparing with Weibo dataset.
The reason is that the text style in the two datasets is formal and the OOV words are less common than Weibo dataset.
In addition, the sequence translation baselines of Moses PB also gained decent results without pre-training or any external data.

According to all experimental results, our approaches still have gaps with the state-of-the-art methods.
Considering the good performance (F1-score 97.7) on the pseudo testing data, the sequence-to-sequence CWS model has shown its capacity on this task and the data scale may be one of main limitations for enhancing our model.

\begin{table*}[t]
  \caption{Experimental results on the CWS benchmark datasets of PKU and MSRA.}
  \label{tab:pkumsra}
  \centering
  \renewcommand\arraystretch{1.1}
  \begin{tabular}{|c|m{5.4cm}<{\centering}|m{0.72cm}<{\centering}|m{0.72cm}<{\centering}|m{0.72cm}<{\centering}|m{0.72cm}<{\centering}|m{0.72cm}<{\centering}|m{0.72cm}<{\centering}|}
  \hline
  \multirow{2}{*}{Groups} &
  \multirow{2}{*}{Models} &
  \multicolumn{3}{c|}{PKU} &
  \multicolumn{3}{c|}{MSRA} \\
  \cline{3-8}
  & &P &R &F &P &R &F \\
  \hline
  A &LTP~\cite{che2010ltp} &95.9 &94.7 &95.3 &86.8 &89.9 &88.3 \\
  \hline \hline
  \multirow{3}{*}{B}
  &Zheng et al., 2013~\cite{zheng2013deep} &93.5 &92.2 &92.8 &94.2 &93.7 &93.9\\ \cline{2-8}
  &Pei et al., 2014~\cite{pei2014max} &94.4 &93.6 &94.0 &95.2 &94.6 &94.9\\ \cline{2-8}
  &Chen et al., 2015~\cite{chen2015gated} &96.3 &95.9 &96.1 &96.2 &96.3 &96.2\\ \cline{2-8}
  &Chen et al., 2015~\cite{chen2015long}  &96.3 &95.6 &96.0 &96.7 &96.5 &96.6\\ \cline{2-8}
  &Cai and Zhao, 2016~\cite{cai2016neural} &95.8 &95.2 &95.5 &96.3 &96.8 &96.5\\ \cline{2-8}
  \hline \hline
  C
  &Zhang et al., 2013~\cite{zhang2013exploring} &- &- &96.1 &- &- &97.4\\ \cline{2-8}
  \hline \hline
  \multirow{7}{*}{M}
  &Moses PB w/ 3-gram LM &92.9 &93.0 &93.0 &96.0 &96.2 &96.1\\ \cline{2-8}
  &Moses PB w/ 5-gram LM &92.7 &92.8 &92.7 &95.9 &96.3 &96.1\\ \cline{2-8}
  &Moses PB w/ 3-gram LM w/ CSC &92.9 &93.0 &92.9 &95.3 &96.5 &95.9 \\ \cline{2-8}
  &Moses PB w/ 5-gram LM w/ CSC &92.6 &93.2 &92.9 &95.9 &96.3 &96.1 \\ \cline{2-8}
  &RNNsearch w/o fine-tuning &93.1 &92.7 &92.9 &84.1 &87.9 &86.0\\ \cline{2-8}
  &RNNsearch  &94.7 &95.3 &95.0 &96.2 &96.0 &96.1 \\ \cline{2-8}
  &RNNsearch w/ post-editing &94.9  &95.4 &95.1 &96.3 &96.1 &96.2 \\ \cline{2-8}
  &RNNsearch w/ CSC &95.2 &94.6 &94.9 &96.1 &96.1 &96.1 \\ \cline{2-8}
  &RNNsearch w/ CSC and post-editing &95.3 &94.7 &95.0 &96.2 &96.1 &96.2 \\ \cline{2-8}
  \hline
  \end{tabular}
\end{table*}

\subsection{Learning CWS and Chinese Spelling Correction Jointly}

As a sequence translation framework, the model can achieve any expected kinds of sequence-to-sequence transformation with the reasonable training.
It hence leaves a lot of space to tackle other NLP tasks jointly.

In this paper, we apply the model to jointly learning CWS and Chinese spelling correction (CSC).
To evaluate the performance of spelling correction, we use automatic method to build two datasets, modified PKU and MSRA,
based on assumptions that
(i) most spelling errors are common with fixed pattern and
(ii) the appearance of spelling errors are randomly.
The details are as follows:
firstly, we construct a correct-to-wrong word pair dictionary counting from
the Chinese spelling check training dataset of SIGHAN 2014~\cite{yu2014overview} as a fixed pattern of spelling errors;
secondly, we randomly select 50\% sentences from PKU and MSRA training set respectively and
replace one of the correct words with the wrong one according to the dictionary for each selected sentence.
The testing set is generated in the same way.

\begin{table}[t]
  \caption{An example of modified data. The character with double underline is wrong and the characters with single underlines are correct.}
  \label{tab:modifieddata}
  \centering
  \renewcommand\arraystretch{1.2}
  \begin{tabular}{|p{2.0cm}<{\raggedleft} |p{9cm}<{\raggedright} |}
  \hline
  Original input &\begin{CJK*}{UTF8}{gbsn}在这个基础上，公安机关还从\uline{\textcolor{blue}{原}}料采购等方面加以严格控制，统一发放“准购证”。\end{CJK*} \\ \hline
  Modified input &\begin{CJK*}{UTF8}{gbsn}在这个基础上，公安机关还从\uuline{\textcolor{red}{源}}料采购等方面加以严格控制，统一发放“准购证”。\end{CJK*} \\ \hline
  Gold standard &\begin{CJK*}{UTF8}{gbsn}在 这个 基础 上 ， 公安 机关 还 从 \uline{\textcolor{blue}{原}}料 采购 等 方面 加以 严格 控制 ， 统一 发放  “ 准 购 证 ” 。\end{CJK*}\\\hline
  \end{tabular}
\end{table}

\begin{table}[t]
  \caption{Experimental results on modified PKU data. The numbers in parentheses represent the changes compared with the normal CWS results shown in Table~\ref{tab:pkumsra}.}
  \label{tab:cscpku}
  \centering
  \begin{tabular}{|m{5.9 cm}<{\centering} |m{1.0cm}<{\centering} |m{1.0cm}<{\centering} |m{1.8cm}<{\centering}|}
  \hline
  Models &P&R&F\\ \hline
  %Modified training data &98.8 &98.6 &98.6 \\ \hline
  %Modified development data &99.6 &99.6 &99.6 \\ \hline
  Modified testing data &99.0 &99.0 &99.0 (-1.0) \\ \hline
  LTP~\cite{che2010ltp} &94.0 &93.2 &93.6 (-1.7)\\ \hline
  Moses PB w/ 3-gram LM &90.8 &91.5 &91.2 (-1.8) \\ \hline
  Moses PB w/ 3-gram LM w/ CSC &92.7 &92.9 &92.8 (-0.1)\\ \hline
  Moses PB w/ 5-gram LM &90.6 &91.3 &91.0 (-1.7) \\ \hline
  Moses PB w/ 5-gram LM w/ CSC &92.3 &93.0  &92.6 (-0.3)\\ \hline
  RNNsearch &93.2 &93.2 &93.2 (-1.8) \\ \hline
  RNNsearch w/ CSC &\textbf{95.0} &\textbf{94.5} &\textbf{94.8} (-0.1) \\ \hline
  \end{tabular}
\end{table}

\begin{table}[!t]
  \caption{Experimental results on modified MSRA data. The numbers in parentheses represent the changes compared with the normal CWS results shown in Table~\ref{tab:pkumsra}.}
  \label{tab:cscmsra}
  \centering
  \begin{tabular}{|m{5.9 cm}<{\centering} |m{1.0 cm}<{\centering} |m{1.0 cm}<{\centering} |m{1.8cm}<{\centering}|}
  \hline
  Models &P&R&F\\ \hline
  Modified testing data &98.5 &98.5 &98.5 (-1.5)\\ \hline
  LTP~\cite{che2010ltp} &84.8 &88.4 &86.6 (-1.7)\\ \hline
  Moses PB w/ 3-gram LM &93.7 &94.6 &94.2 (-1.9) \\ \hline
  Moses PB w/ 3-gram LM w/ CSC &95.0  &96.3 &95.6 (-0.3)\\ \hline
  Moses PB w/ 5-gram LM &93.7 &94.7 &94.2 (-1.9)\\ \hline
  Moses PB w/ 5-gram LM w/ CSC &94.6  &95.9 &95.3 (-0.7)\\ \hline
  RNNsearch &93.8 &94.7 &94.2 (-1.9) \\ \hline
  RNNsearch w/ CSC &\textbf{96.0} &\textbf{96.0} &\textbf{96.0} (-0.1) \\ \hline
  \end{tabular}
\end{table}

We treat the modified sentences and the original segmented sentences as
the input sequence and the golden standard respectively in the training procedure.
Table~\ref{tab:modifieddata} gives an example of the modified data. In the testing procedure,
we send the sentence with wrong words into the model,
and expect to get the segmented sentence with all correct words.
The results are shown in Table~\ref{tab:cscpku} and Table~\ref{tab:cscmsra}.
Since the general CWS tool LTP does not have the ability to correct spelling mistakes, the performance decreases.
Whereas, the impact of the wrong words is limited in our models trained to do CWS and CSC jointly.

\section{Related Work}

CWS using neural networks have gained much attention in recent years as
they are capable of learning features automatically.
Collobert et al.~\cite{collobert2011natural} developed a general neural architecture for sequence labelling tasks.
Zheng et al.~\cite{zheng2013deep} adapted word embedding and
the neural sequence labelling architecture~\cite{collobert2011natural} for CWS.
Pei et al.~\cite{pei2014max} improved upon Zheng et al.~\cite{zheng2013deep} by
modeling complicated interactions between tags and context characters.
Chen et al.~\cite{chen2015gated} proposed gated recursive neural networks to model the combinations of context characters.
Chen et al.~\cite{chen2015long} introduced LSTM into neural CWS models to capture the potential long-distance dependencies.
However, the methods above all regard CWS as sequence labelling with local input features.
Cai and Zhao~\cite{cai2016neural} re-formalize CWS as a direct segmentation learning task without the above constrains,
but the maximum length of words is limited.

\textbf{ Sequence-to-sequence Machine Translation Models. }
Neural sequence-to-sequence machine translation models have rapid developments since 2014.
Cho et al.~\cite{cho2014learning} proposed an RNN encoder-decoder framework with
gated recurrent unit to learn phrase representations.
Sutskever et al.~\cite{sutskever2014sequence} applied LSTM for RNN encoder-decoder framework to
establish a sequence-to-sequence translation framework.
Bahdanau et al.~\cite{bahdanau2014neural} improved upon Sutskever et al.~\cite{sutskever2014sequence} by
introducing an attention mechanism.
Wu et al~\cite{wu2016google} presented Google's Neural Machine Translation system
which is serving as an online machine translation system.
Gehring et al.~\cite{gehring2017convs2s} introduce an architecture based entirely on convolutional neural networks to
sequence-to-sequence learning tasks which improved translation accuracy at an order of magnitude faster speed.
Other efficient sequence-to-sequence models will be introduced into this task and compared with existing works in our future work.

\section{Conclusion}

In this paper, we re-formalize the CWS as a sequence-to-sequence translation problem and
apply an attention based encoder-decoder model.
We also make an attempt to let the model jointly learn CWS and CSC.
Furthermore, we propose an LCS based post-editing algorithm to deal with potential translating errors.
Experimental results show that our approach achieves competitive performances compared with
state-of-the-art methods both on normal CWS and CWS with CSC.

In the future, we plan to apply other efficient sequence-to-sequence models for CWS and study an end-to-end framework for multiple natural language pre-processing tasks.

\subsubsection*{Acknowledgments}
This work was supported by the National Basic Research Program (973) of China (No. 2013CB329303) and
the National Natural Science Foundation of China (No. 61132009).

% ---- Bibliography ----
\bibliography{37_references}
\bibliographystyle{splncs04}

\end{document}